\pdfoutput=1

\documentclass[11pt]{article}

\usepackage[]{EMNLP2023}

\usepackage{times}
\usepackage{latexsym}
\usepackage{amsmath}
\usepackage{amsfonts}

\usepackage[T1]{fontenc}

\usepackage[utf8]{inputenc}

\usepackage{microtype}

\usepackage{inconsolata}

\usepackage{braket}
\usepackage{verbatim}
\usepackage{ulem}
\usepackage{adjustbox}
\usepackage{multirow}
\usepackage{enumitem}

\usepackage[percent]{overpic} 
\usepackage{xcolor}
\usepackage{hyperref}

\title{Unlearning in LLMs: Methods, Evaluation, and Open Challenges}

\author{ Tyler Lizzo, Larry Heck\\ AI Virtual Assistant (AVA) Lab\\
   Georgia Institute of Technology  \\ \texttt{\{lizzo,larryheck\}@gatech.edu}}

\begin{document}
\maketitle
\setlength{\belowcaptionskip}{-10pt}

\begin{abstract}
Large language models (LLMs) have achieved remarkable success across natural language processing tasks, yet their widespread deployment raises pressing concerns around privacy, copyright, security, and bias. Machine unlearning has emerged as a promising paradigm for selectively removing knowledge or data from trained models without full retraining. In this survey, we provide a structured overview of unlearning methods for LLMs, categorizing existing approaches into data-centric, parameter-centric, architecture-centric, hybrid, and other strategies. We also review the evaluation ecosystem, including benchmarks, metrics, and datasets designed to measure forgetting effectiveness, knowledge retention, and robustness. Finally, we outline key challenges and open problems, such as scalable efficiency, formal guarantees, cross-language and multimodal unlearning, and robustness against adversarial relearning. By synthesizing current progress and highlighting open directions, this paper aims to serve as a roadmap for developing reliable and responsible unlearning techniques in large language models.

\end{abstract}

\begin{figure*}
  \centering
  \begin{overpic}[width=\linewidth]{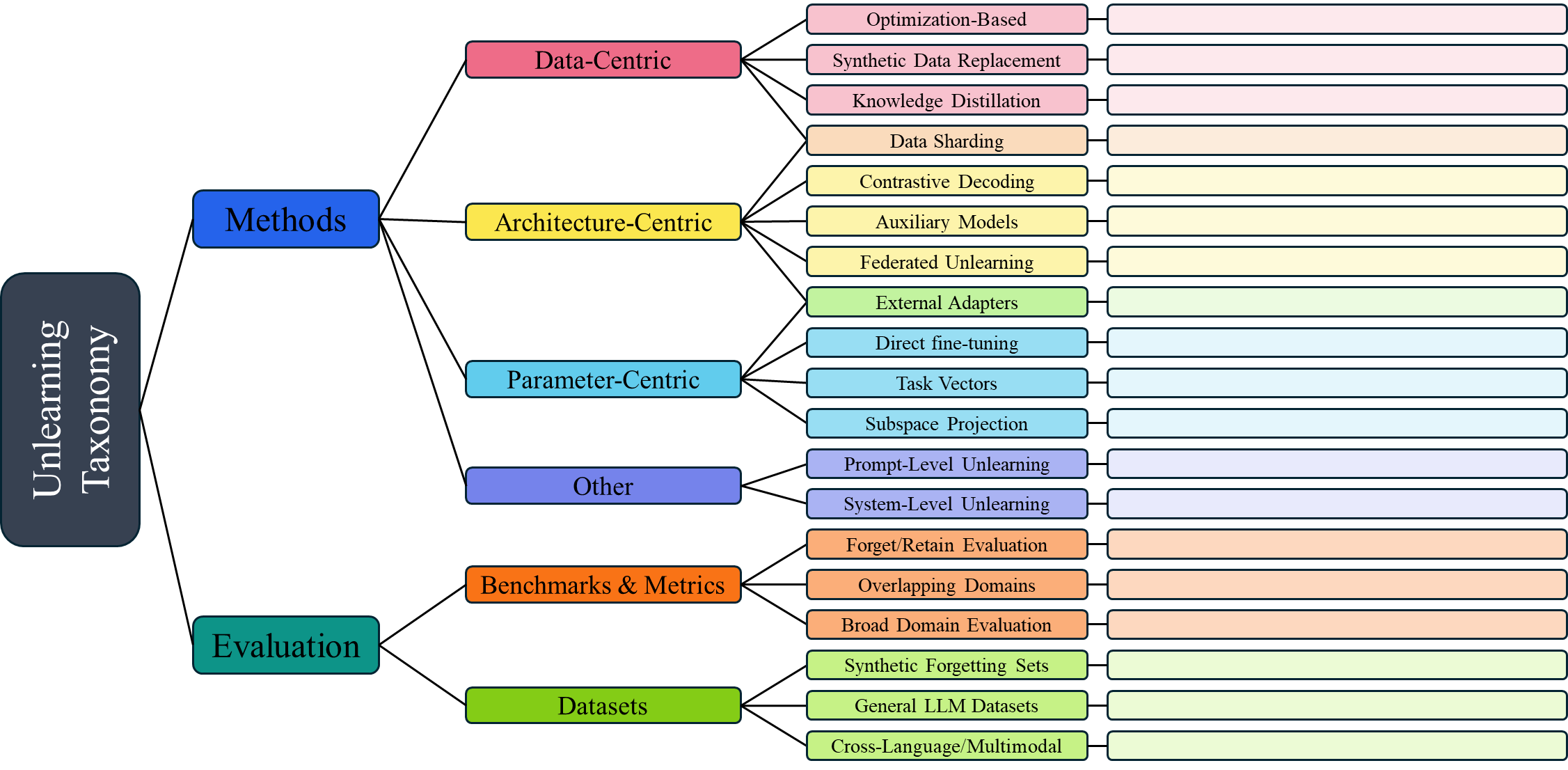} 
    \put(71,47.6){\tiny \citet{jang2023knowledge,liu2022continual,wang2025balancing}}
    \put(71,45.022222){\tiny \citet{chen-etal-2025-ynu-semeval,zhu2025llmunlearningexpertcurated}}
    \put(71,42.444444444){\tiny \citet{wang2023kgageneralmachineunlearning}; \citet{wang2024efficienteffectiveunlearninglarge}}
    \put(71,39.8666666){\tiny \citet{bourtoule2021machineunlearning,kadhe2023fairsisaensemblepostprocessingimprove}}
    \put(71,37.288888888){\tiny \citet{ji2024reversingforgetretainobjectivesefficient,suriyakumar2025ucdunlearningllmscontrastive}}
    \put(71,34.711111111){\tiny \citet{mitchell2022memorybasedmodeleditingscale,sanyal2025agentsneedllmunlearning}
}
    \put(71,32.133333){\tiny \citet{liu2021federatedunlearning,zhang2025oblivionislightweightlearningunlearning}}
    \put(71,29.555555){\tiny \citet{chen2023unlearnwantforgetefficient,ding2025unifiedparameterefficientunlearningllms}
}
    \put(71,26.977777){\tiny \citet{eldan2023whosharrypotterapproximate,gao2025largelanguagemodelcontinual}}
    \put(71,24.4){\tiny \citet{ilharco2023editing,gao2024ethos}}
    \put(71,21.8222222){\tiny \citet{lizzo2025unlearn}}
    \put(71,19.244444){\tiny \citet{thaker2024guardrailbaselinesunlearningllms,pawelczyk2024incontextunlearninglanguagemodels}}
    \put(71,16.666666){\tiny \citet{wang2024machineunlearningmeetsretrievalaugmented}}
    \put(71,14.0888888){\tiny \citet{maini2024tofutaskfictitiousunlearning,li2024wmdpbenchmarkmeasuringreducing}}
    \put(71,11.511111){\tiny \citet{hu2025blurbenchmarkllmunlearning}}
    \put(71,8.933333){\tiny \citet{lizzo2025unlearn}}
    \put(71,6.355555){\tiny \citet{maini2024tofutaskfictitiousunlearning,li2024wmdpbenchmarkmeasuringreducing}}
    \put(71,3.777777){\tiny \citet{srivastava2023imitationgamequantifyingextrapolating,liang2023holisticevaluationlanguagemodels}}
    \put(71,1.2){\tiny \citet{lu2025learnunlearnaddressingmisinformation,patil2025unlearningsensitiveinformationmultimodal}}
    \put(29.7,44.5){ \hyperref[datacentric]{\phantom{XXXXXXXXXI}}}
    \put(29.7,34.1){ \hyperref[architecturecentric]{\phantom{XXXXXXXXXI}}}
    \put(29.7,24.1){ \hyperref[parametercentric]{\phantom{XXXXXXXXXI}}}
    \put(29.7,17.3){ \hyperref[otherapproaches]{\phantom{XXXXXXXXXI}}}
    \put(29.7,11){ \hyperref[benchmarks]{\phantom{XXXXXXXXXI}}}
    \put(29.7,3.3){ \hyperref[datasets]{\phantom{XXXXXXXXXI}}}

  \end{overpic}
  \caption{Taxonomy of LLM Unlearning Methods and Evaluation Techniques}
\end{figure*}
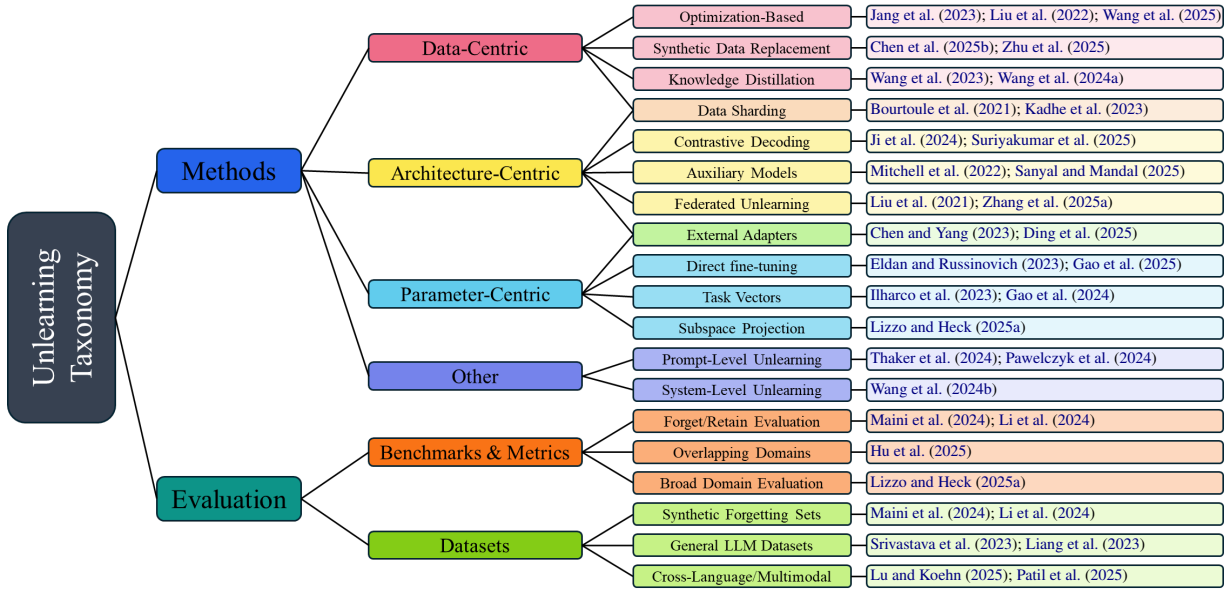

\section{Introduction}

In recent years, large language models (LLMs) have transitioned from research artifacts to widely deployed systems supporting millions of users across diverse domains. With this scale of adoption, their limits have become more visible. While LLMs are impressively general, they often lag behind in domains like arithmetic, counterfactual reasoning, and complex causal inference—areas where specialized modules tend to outperform them in both correctness and computational cost. Recent work in tool-augmented LLMs  \cite{Chen2024ToolAugmentedLLMs} demonstrates the value of routing reasoning and API calls to external components rather than forcing the model itself to execute all reasoning internally. This line of research prompts a broader question: might overall efficiency and robustness improve if LLMs could judiciously shed redundant or undesirable internal representations, instead delegating to external tools or modules when appropriate?

The case for unlearning is not limited to efficiency. Regulatory requirements such as GDPR, copyright disputes, and broader ethical concerns all highlight the need to remove information or behaviors after a model has been trained. Beyond compliance, unlearning also creates opportunities to reduce biased reasoning, eliminate toxic responses, and guard against adversarial backdoors. Unlike retraining or fine-tuning, which adjust model behavior globally, unlearning is designed to be targeted and to remove specific information while leaving overall performance intact.

This survey synthesizes the growing body of research on unlearning in LLMs. We organize existing approaches into four categories: data-centric, parameter-centric, architecture-centric, and others. We also review the benchmarks, datasets, and evaluation metrics that have been proposed. Ongoing challenges include providing guarantees of forgetting, scaling methods to billion or trillion--parameter models, and addressing multilingual or cross-lingual cases. By situating unlearning within the wider machine learning landscape, including continual learning, model editing, and alignment, we aim to map out current progress and highlight open directions for building LLMs that are both efficient and socially responsible.

\section{History of Unlearning}

Machine unlearning predates its application to large language models and was initially motivated by privacy regulations and the need to remove specific training examples from deployed systems. Early work concentrated on exact unlearning, where the goal was to make a model trained on a reduced dataset indistinguishable from one retrained from scratch. Techniques such as SISA training \citep{bourtoule2021machineunlearning}, which partitioned data into shards to localize retraining, and influence-function-based methods, which estimated the effect of removing a single data point, provided the first formalizations of the problem. Other directions explored the use of differential privacy to bound the contribution of individual data points to the parameters.

While these approaches were conceptually important, they generally break down at modern scale. Retraining or even partial retraining is prohibitively expensive for billion or trillion--parameter models. Methods that require specialized training pipelines, such as checkpointing or data sharding, are also impractical once a model has already been fully trained and deployed.

It is also useful to separate unlearning from neighboring ideas. Retraining from scratch trivially eliminates unwanted data but is computationally infeasible in most cases. Model pruning removes parameters to compress networks, but it does not selectively target knowledge. Fine-tuning may suppress certain behaviors but often introduces catastrophic forgetting of unrelated abilities. Machine unlearning is instead defined by its focus on selective removal of information while maintaining general utility. This makes it uniquely challenging to implement, but also uniquely valuable in the context of large language models.
\section{Taxonomy of Approaches}
Unlearning in large language models has emerged through a variety of methodological perspectives, each focusing on a different locus of intervention within the model lifecycle. To provide structure to this rapidly evolving field, we categorize existing approaches into four broad classes: (\ref{datacentric}) \textit{data-centric},  (\ref{architecturecentric}) \textit{architecture-centric}, (\ref{parametercentric}) \textit{parameter-centric}, (\ref{hybrid}) \textit{hybrid} methods, and (\ref{otherapproaches}) other methods. 

Data-centric approaches operate primarily at the level of training data and optimization, aiming to mimic the effect of retraining from scratch on a filtered dataset by techniques such as selective retraining, gradient ascent, knowledge distillation, or the use of synthetic replacement data. Parameter-centric methods intervene directly on the weights of a pre-trained model, ranging from full fine-tuning with specialized objectives to parameter-efficient updates (e.g., task vectors, adapter subtraction), as well as localized weight modifications and subspace-based methods. Architecture-centric approaches modify or extend the model structure itself, introducing new modules (e.g., adapter layers) or transformation layers designed to isolate and suppress specific knowledge, often with the goal of efficiency and modularity. Finally, hybrid methods combine elements across categories, such as ensemble–student distillation pipelines, prompt-based interventions, or editing–unlearning hybrids that bridge the gap between factual editing and targeted forgetting. 

This taxonomy not only clarifies the design space of current unlearning techniques, but also highlights the trade-offs each class embodies in terms of forgetting guarantees, efficiency, and preservation of non-targeted knowledge. In the subsections that follow, we review each category in detail, summarizing representative methods, their mechanisms, and limitations.

\subsection{Data-Centric Unlearning}\label{datacentric}

Data-centric approaches to unlearning operate directly at the level of training data and optimization. The central idea is to alter the learning signal such that the model’s behavior on the forget set mimics that of a model trained without the data, while retaining utility on the rest of the corpus. Compared to parameter- or architecture-centric methods, these approaches treat unlearning primarily as a variant of retraining, modifying the data or objectives rather than localizing parameters.

\paragraph{Optimization-Based Unlearning.}\label{gradientbased}
\citet{jang2023knowledge} took the gradient descent training approach and flipped it to maximize loss instead of minimizing it:
\begin{footnotesize}
\begin{align*}
\mathcal{L}_{UL}(f_\theta, \mathbf{x}) = - \sum_{t=1}^T \log\Bigl(p_\theta(x_t|x_{<t})\Bigr)
\end{align*}
\end{footnotesize}
thereby updating the model in the direction opposite to the training gradient.  
Such gradient-ascent approaches push the parameters away from representations of unwanted data, but they are notoriously unstable at scale and often cause collateral degradation in unrelated capabilities.  
Subsequent work sought to stabilize this process by adding corrective gradient-descent steps \citep{liu2022continual} or by using KL-divergence objectives on a retain set \citep{wang2025balancing}. Forget-data-only Loss AdjustmenT (FLAT; \citealp{wang2024flat}) took this further by replacing the inverted loss with a principled $f$-divergence maximization between template responses and responses to the forget set.

Preference-based objectives provide another stabilization mechanism. Preference Optimization (PO; \citealp{rafailov2024directpreferenceoptimizationlanguage}) and its unlearning variant, Negative Preference Optimization (NPO; \citealp{zhang2024npo}), recast forgetting as a preference-alignment problem in which the model is encouraged to assign lower preference to responses associated with the forget set. NPO reduces to gradient ascent in the high-temperature limit, but its preference-informed objective significantly slows the collapse dynamics characteristic of GA, yielding more stable optimization and improved trade-offs between unlearning effectiveness and overall model utility.

\paragraph{Synthetic Data Replacement.}\label{syntheticdata}
Another line of data-centric work replaces the forget set with carefully constructed synthetic samples during fine-tuning to weaken associations with the original inputs. \citet{zhu2025llmunlearningexpertcurated} generates domain-specific synthetic forget sets using LLMs, showing performance close to expert-curated data in areas such as biosecurity and fiction.  A related approach, Synthetic Token Alternative Training (STAT) \citep{chen-etal-2025-ynu-semeval}, introduces pseudo tokens for the forget set and alternates optimization between augmented forget samples and retain data, helping to stablizing unlearning. While these approaches avoid full retraining and offer flexibility, their success depends heavily on the quality of synthetic data and may be challenging to scale to complex domains.

\paragraph{Knowledge Distillation.}\label{knowledgedistillation}
Within the data-centric paradigm, another direction frames unlearning as a student–teacher problem. Instead of directly applying gradient updates on the forget set, the model is retrained to align its behavior with a teacher model that either has not seen the forget data or has been explicitly designed to suppress it. The Knowledge Gap Alignment (KGA) method \citep{wang2023kgageneralmachineunlearning} exemplifies this approach by minimizing the divergence between student predictions on the forget set and teacher predictions on unrelated data, thereby aligning the “knowledge gap” with behavior consistent to a model never trained on the forget data. More recently, E2URec \citep{wang2024efficienteffectiveunlearninglarge} improves both efficiency and effectiveness by combining distillation with parameter-efficient fine-tuning: only a small number of LoRA parameters are updated, while multiple teacher models are maintained to guide the student during unlearning. Distillation-based approaches offer flexibility across model architectures and tasks, but typically require auxiliary teachers and access to external data, making them costly at scale. Their effectiveness further depends on the size of the forget set and the diversity of external data, which can limit applicability in real-world LLM deployments.

\begin{figure}
    \centering
    \includegraphics[width=0.9\linewidth]{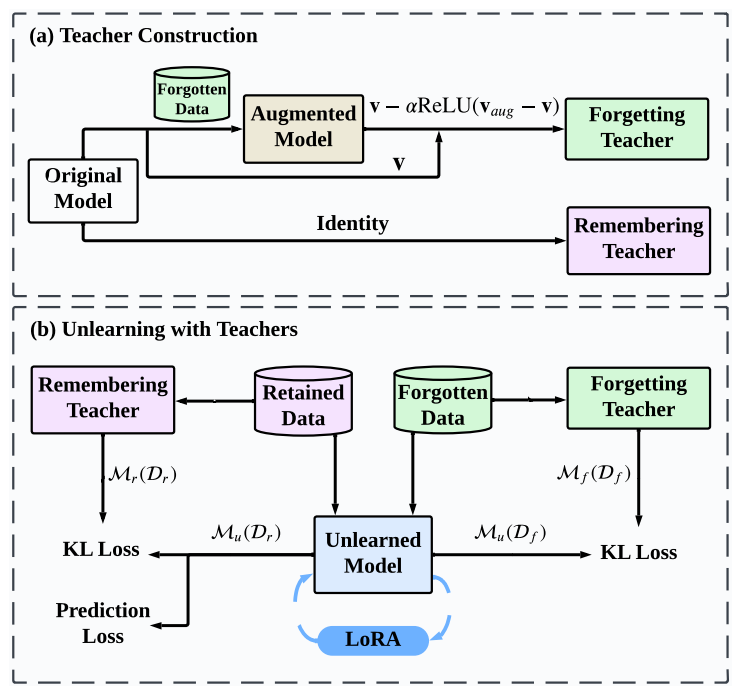}
    \caption{The teacher-student paradigm proposed in E2URec \cite{wang2024efficienteffectiveunlearninglarge}, showing both the construction of the teachers and how the student model uses those teachers.}
    \label{fig:placeholder}
\end{figure}

Overall, data-centric methods offer conceptually clear pathways for unlearning and provide a useful foundation for evaluation. Yet their reliance on retraining-like procedures limits their practicality for large language models, where efficiency and scalability are paramount.
\subsection{Architecture-Centric Unlearning}\label{architecturecentric}
Architecture-centric unlearning methods modify or extend the structure of the model or its training system, rather than focusing exclusively on data or parameter updates. These approaches often aim to encapsulate or isolate knowledge in modular components, making it easier to remove or suppress unwanted behavior without retraining the entire network.

\begin{figure}
    \centering
    \includegraphics[width=0.9\linewidth]{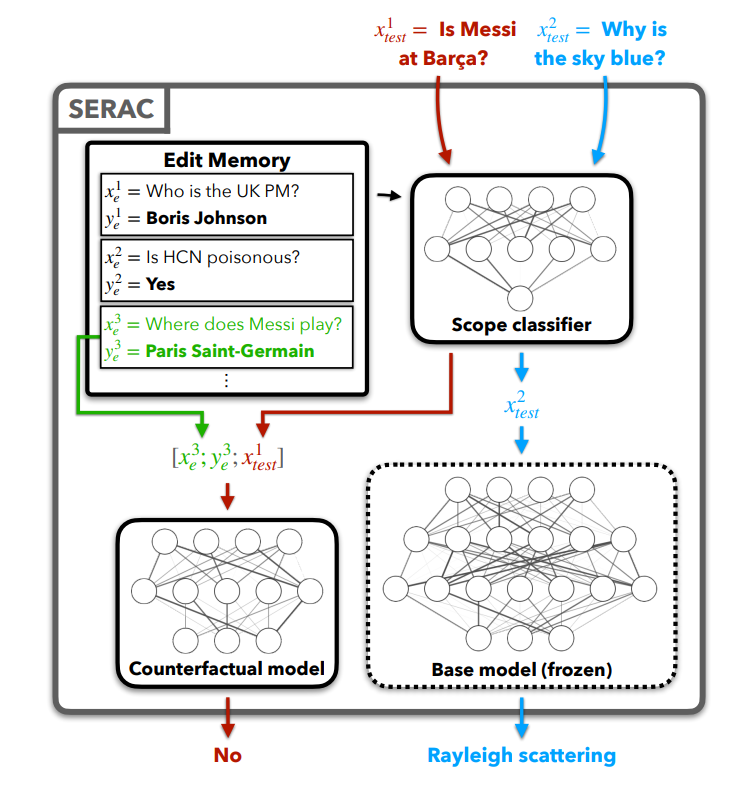}
    \caption{Semi-Parametric Editing with a Retrieval-Augmented Counterfactual Model (SERAC) \cite{mitchell2022memorybasedmodeleditingscale}}
    \label{fig:serac}
\end{figure}

\paragraph{Auxiliary Models.} 
Another strategy is to attach auxiliary modules, such as specialized classifiers or agents, that suppress forgotten content while not touching the base model. This modularity provides efficiency, as only small components are trained or modified. One work connects unlearning with model editing, where auxiliary memory and retrieval modules localize edits, shown in Fig. \ref{fig:serac} \citep{mitchell2022memorybasedmodeleditingscale}. This work stores edits in an explicit memory and reasons over them via a counterfactual model, enabling precise suppresion of knowledge. Another recent work, Agentic LLM Unlearning (ALU) \citep{sanyal2025agentsneedllmunlearning}, leverages multiple LLM agents with each dedicated to a step of the unlearning process. While auxiliary models offer efficiency and reversibility, they leave the underlying parameters intact. Contrastive decoding is a specialized form of this paradigm, but its reliance on auxiliary models to guide inference warrants its dedicated discussion.

\paragraph{Contrastive Decoding.}  
One promising direction is to intervene at the decoding stage rather than altering model weights. Contrastive decoding introduces additional models to guide generation such that outputs relying on the forget set are down-weighted. A representative example is Unlearning from Logit Difference (ULD) \citep{ji2024reversingforgetretainobjectivesefficient}, which trains an assistant LLM on the forget set and then derives the unlearned model by subtracting its logits from those of the target model. The assistant is optimized with a reversed objective, remembering the forget data while suppressing retain knowledge:  
\begin{footnotesize}
\begin{align*}
\min_{\phi} \mathcal{L}(\phi)= \min_\phi \mathcal{L}_f(\phi) - \beta L_r(\phi),
\end{align*}
\end{footnotesize}

where $\mathcal{L}_f$ and $\mathcal{L}_r$ denote the forget and retain losses. Other recent work extends the idea by training two smaller auxiliary models—one with and one without the forget set—and using their logit difference to steer the original LLM at inference time \citep{suriyakumar2025ucdunlearningllmscontrastive}. These approaches substantially improve the trade-off between forgetting effectiveness and knowledge retention, with results on benchmarks such as TOFU and MUSE showing strong utility preservation. While contrastive decoding is attractive for its reversibility and lack of parameter modification, it requires maintaining multiple model variants in memory and can be computationally intensive during inference.

\begin{figure}
    \centering
    \includegraphics[width=0.9\linewidth]{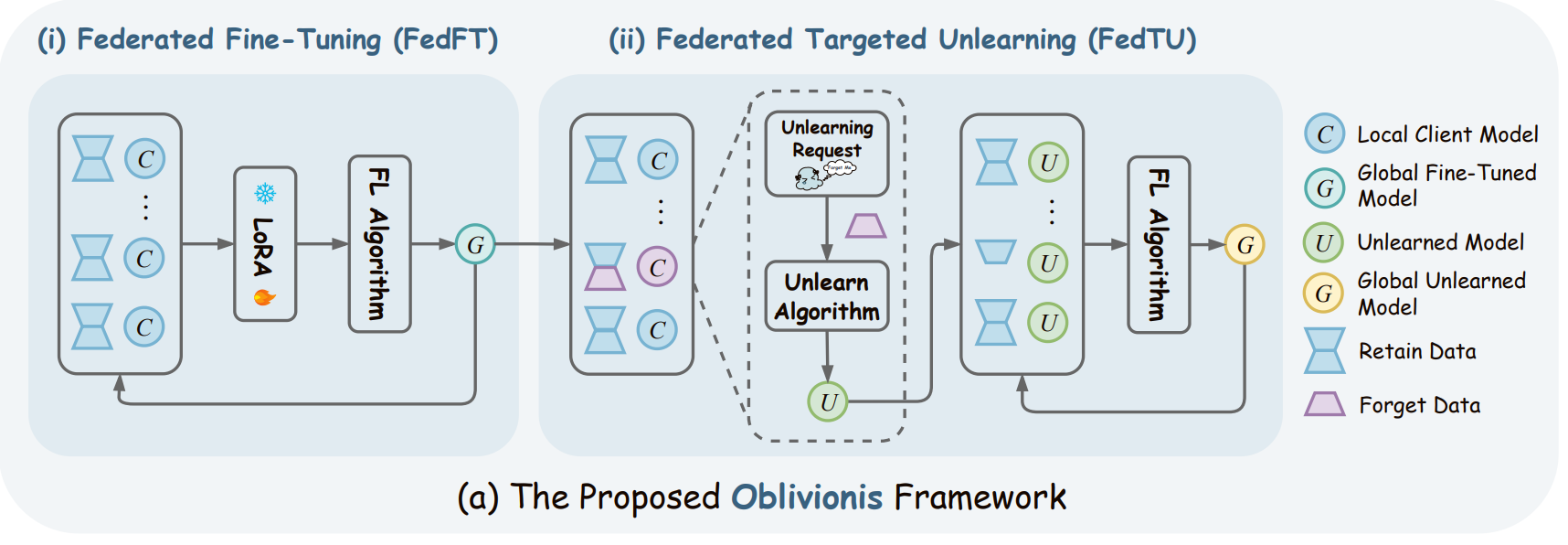}
    \caption{The Oblivionis Federated Unlearning Pipeline}
    \label{fig:oblivionis}
\end{figure}

\paragraph{Federated Unlearning.}  
In federated learning (FL) contexts, unlearning is defined at the architectural level: specific client updates are removed from the global model with modified aggregation steps to negate contributions from clients with forget requests. The earliest approach, FedEraser \citep{liu2021federatedunlearning}, reconstructs an unlearned global model by leveraging stored historical updates from each client. By calibrating these retained updates and using them in place of retraining, FedEraser eliminates the influence of the forgotten client while significantly reducing reconstruction time compared to training from scratch. Due to regulatory requirements, there have been a lot of recent extensions to Federated Unlearning, including the Oblivionis framework \citep{zhang2025oblivionislightweightlearningunlearning} shown in Fig. \ref{fig:oblivionis}, which integrates a number of FL and unlearning algorithms into a unified pipeline to better demonstrate the interplay between the two. Federated unlearning is thus crucial for privacy-sensitive distributed deployments, but it faces unique challenges with distributed data, stringent privacy constraints, and interdependent aggregation \cite{bhansali2024legolanguagemodelbuilding}.

Overall, architecture-centric methods emphasize modularity and system-level flexibility. They often provide efficient or reversible pathways for unlearning, but may struggle to guarantee complete removal of targeted information, especially in settings where adversarial extraction is a concern.
\subsection{Parameter-Centric Unlearning}\label{parametercentric}
Parameter-centric methods target the weights of a pre-trained model directly. Rather than modifying training data or introducing auxiliary structures, these approaches identify and manipulate parameter updates to erase targeted knowledge while minimizing collateral damage. They are attractive for large language models because they often avoid full retraining and can be implemented efficiently with modern parameter-efficient fine-tuning techniques.

\paragraph{Direct Fine-Tuning.}  
A straightforward parameter-centric strategy is to fine-tune model weights using objectives designed to suppress the forget set. Full fine-tuning of all parameters is prohibitively expensive for modern LLMs, making parameter-efficient variants such as Low-Rank Adaptation (LoRA) \citep{hu2021loralowrankadaptationlarge} particularly attractive. LoRA freezes the pre-trained weights and injects small trainable rank-decomposition matrices into each Transformer layer, reducing the number of trainable parameters by several orders of magnitude while maintaining model quality.
\begin{equation*}
    h=W_0x+\Delta Wx =W_0x+BAx
\end{equation*}
where $W_0\in\mathbb{R}^{d\times k}$, $B\in \mathbb{R}^{d\times r}$, $A\in \mathbb{R}^{r\times k}$, and the rank $r \ll \min(d,k)$.
For unlearning, LoRA modules can be trained on the forget set and then negated or removed, effectively canceling their contribution without altering the full weight matrix \citep{gao2025largelanguagemodelcontinual}.  

Recent work further refines fine-tuning by incorporating targeted objectives to erase specific data. One approach uses a reinforced model to identify tokens most related to the unlearning target by comparing its logits to a baseline model, then replaces those expressions with generic counterparts and generates alternative labels for each token \cite{eldan2023whosharrypotterapproximate}. Fine-tuning on these alternative labels effectively erases the original text associations, suppressing recall of the target data when prompted.

\paragraph{Task Vectors.}  
\begin{figure}
    \centering
    \includegraphics[width=0.9\linewidth]{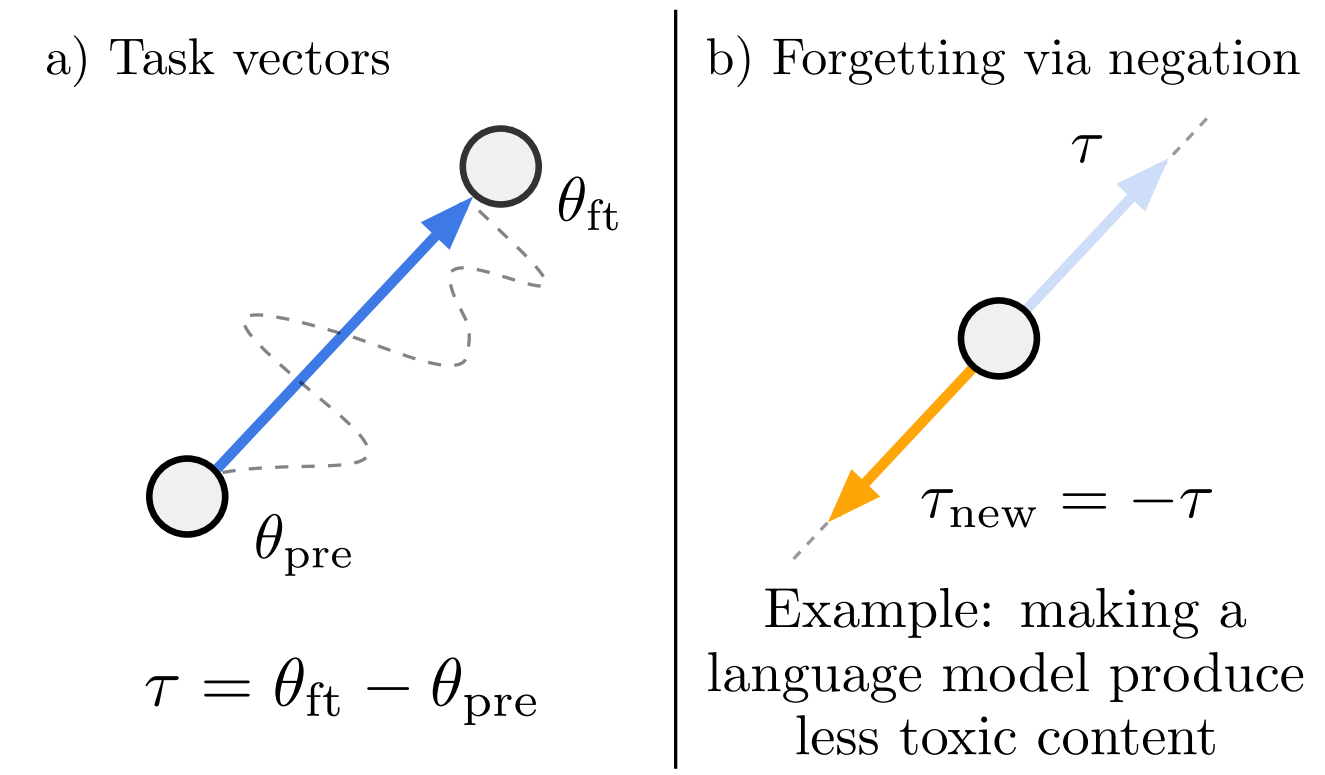}
    \caption{The Task Vector method \cite{ilharco2023editing}}
    \label{fig:placeholder}
\end{figure}

Task vector methods, introduced by \citet{ilharco2023editing}, specify a direction in the weight space of a model to improve or attentuate the performance on a task. It does this via task vectors, defined as $\tau = \theta_{\text{ft}} - \theta_{\text{pre}}$, where $\theta_{\text{ft}}$ and $\theta_{\text{pre}}$ are the fine-tuned and base model parameters, respectively. By adding or subtracting such vectors, one can steer model behavior without additional training. For unlearning, task-vector negation has been used to suppress unwanted behaviors such as toxic or biased generation \citep{liu2024towardssafer,zhou2024making}. Ethos \citep{gao2024ethos} extends task-vector methods by projecting vectors onto principal components via SVD, separating undesired from general knowledge and negating only the harmful components, thereby reducing damage to model utility.  

\begin{figure*}[t]
    \centering
    \includegraphics[width=\linewidth]{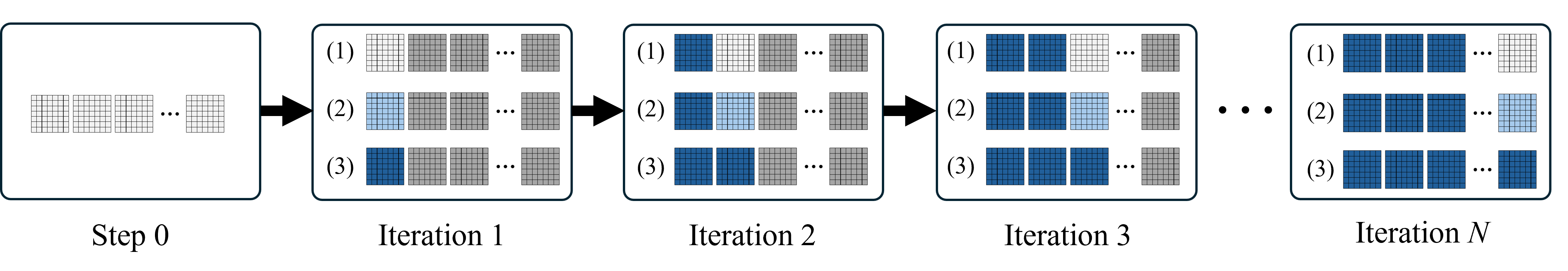}
    \caption{The Subspace Identification Process. The process begins by randomly initializing the model weights and then freezing them. Then an iterative process of unfreezing, training, and refreezing each layer occurs. This results in a set of matrices that capture an accurate representation for that task.}
    \label{fig:identification}
\end{figure*}
\begin{figure}[!b]
    \centering
    \includegraphics[width=0.9\linewidth]{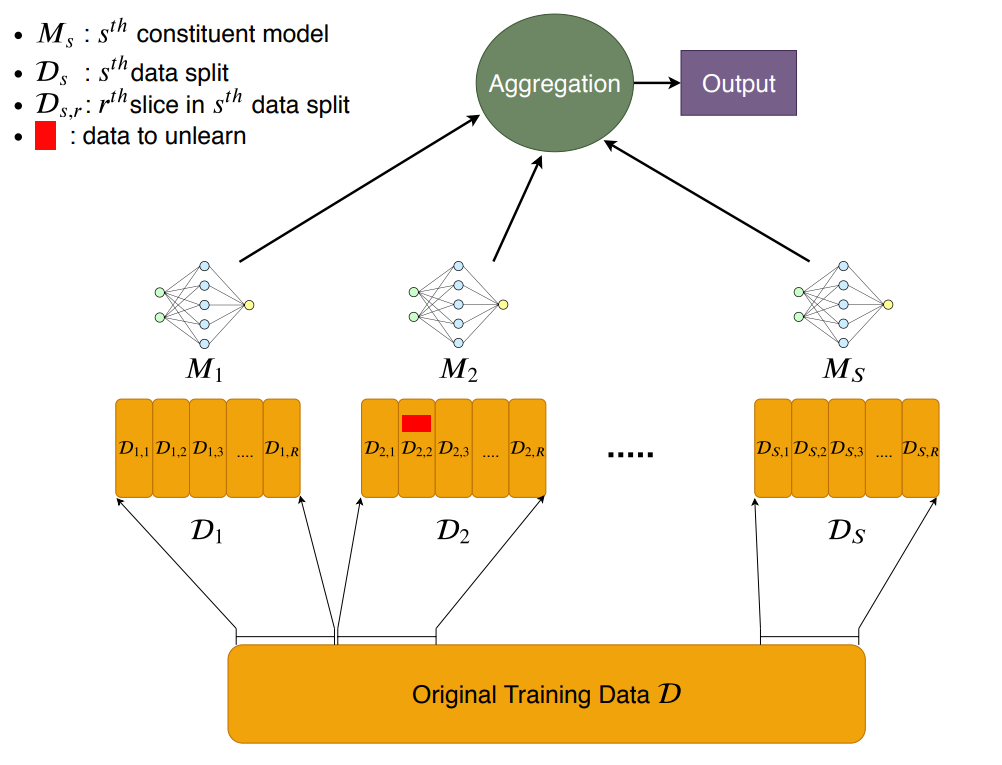}
    \caption{SISA training: data is divided into shards and slices. Models are trained for each shard, and these models are aggregated to get the full model \cite{bourtoule2021machineunlearning}}
    \label{fig:placeholder}
\end{figure}
\paragraph{Subspace Projection.}
Recent work has focused on identifying the low-dimensional subspace corresponding to a task within the weights of an LLM. The \textsc{UNLEARN} method, introduced by \citet{lizzo2025unlearn,lizzo2025patent}, formalizes this idea by sequentially freezing and training layers to isolate a task-specific subspace, $T_i$ (see Fig. \ref{fig:identification}).

A critical innovation in \textsc{UNLEARN} is the introduction of \emph{subspace discrimination}. With two similar tasks, naive subtraction of one can degrade performance on the other. Subspace discrimination mitigates this by explicitly estimating the intersection between the forget subspace and the retain subspace, through a process that combines singular value decomposition with Graham-Schmidt orthogonalization:
\begin{footnotesize}
\begin{align*}
SV_k(T'_i)=SV_k(T_i)-\sum_{j=1}^N\frac{SV_k(T_i)\cdot SV_j(T_o)}{SV_j(T_o)\cdot SV_j(T_o)}SV_j(T_o)
\end{align*}
\end{footnotesize}
This gives a lightweight set of matrices that represent the weight space unique to the task of interest. The full weight space is then projected onto these matrices to remove the task. \textsc{UNLEARN} demonstrates significant performance gains both on selective forgetting and knowledge retention of adjacent tasks.

Overall, parameter-centric methods strike a balance between flexibility and efficiency. They are particularly promising for LLMs, as they align with the trend toward parameter-efficient adaptation. Yet their reliance on weight-space manipulations raises open questions about robustness, guarantees of forgetting, and scalability to complex multi-task scenarios.

\subsection{Hybrid Approaches}\label{hybrid}
While many unlearning methods can be categorized cleanly as data-, parameter-, or architecture-centric, some approaches span multiple dimensions. These hybrid techniques combine the strengths of different perspectives, often for efficiency or modularity, but they also inherit limitations from each side.

\paragraph{Data Sharding.}  
SISA training (Sharded, Isolated, Sliced, and Aggregated) exemplifies the hybrid nature of some unlearning strategies by combining data partitioning with architectural modularity \citep{bourtoule2021machineunlearning}. The dataset is divided into disjoint shards, and separate sub-models or slices are trained independently before being aggregated into a global model. When a forget request arise, only the sub-models with relevant data must be retrained.The original SISA framework demonstrated significant speedups for image classification and tabular datasets by leveraging this partitioned structure \citep{bourtoule2021machineunlearning}. Expansions of this original work demonstrate that, while scalable, it can also negatively impact LLM fairness \citep{kadhe2023fairsisaensemblepostprocessingimprove}. To address this, FairSISA introduces post-processing bias mitigation techniques that adapt fairness interventions without sacrificing the unlearning benefits \citep{kadhe2023fairsisaensemblepostprocessingimprove}. Although conceptually appealing, SISA-style approaches are rarely practical for modern LLMs due to the need for specialized training pipelines and greatly increased storage overhead.

\paragraph{External Adapters.}  
Another hybrid family arises combines architectural modifications with parameter updates by adding external adapters on top of the base model to suppress unwanted knowledge. One early example is the EUL framework proposed by \citep{chen2023unlearnwantforgetefficient}, which adds unlearning layers into Transformer models and fuses them to handle forget requests. LLMEraser \citep{ding2025unifiedparameterefficientunlearningllms} refines this approach by categorizing unlearning tasks and applies precise parameter adjustments to adapters using influence functions.

Overall, hybrid methods demonstrate how different design choices can be combined to achieve flexible unlearning. They are conceptually appealing but often face challenges of practicality and completeness, especially at the scale of modern LLMs.

\subsection{Other Approaches}
\label{otherapproaches}
Finally, some approaches to unlearning do not fit neatly into the data-, parameter-, or architecture-centric families, nor do they combine them in hybrid form. Instead, they operate at different levels of interaction with the LLM. Two notable directions are prompt-level unlearning and system-level unlearning.

\paragraph{Prompt-Level Unlearning.}  
Prompt-level methods intervene at inference time rather than altering model parameters. Guardrails \citep{thaker2024guardrailbaselinesunlearningllms}, prompt rewriting \citep{pawelczyk2024incontextunlearninglanguagemodels}, and instruction-tuned filters \citep{liu2024largelanguagemodelunlearning} can be used to block or redirect outputs related to the forget set. For example, a model may be instructed to respond with ``I cannot answer'' when asked about specific topics. These approaches are lightweight, reversible, and do not require retraining, making them attractive in practice. However, they are fundamentally behavioral rather than representational: the underlying model still encodes the forgotten knowledge, which may be recoverable through adversarial prompting or fine-tuning.

\paragraph{System-Level Unlearning.}  
Another class of methods targets the broader system surrounding the LLM rather than its internals. Retrieval-augmented generation (RAG) allows an LLM to access documents from an external knowledge base and condition its output on this added context \citep{reichman2024densepassageretrievalretrieving}. In this setting, unlearning can be achieved by modifying this knowledge base without altering the base model \citep{wang2024machineunlearningmeetsretrievalaugmented}. Such RAG-based approaches are particularly valuable for closed-source LLMs, where there is no direct parameter access and traditional unlearning methods often fail. While system-level methods are efficient and practical for deployment, their guarantees are limited: they control outputs by shaping access to information rather than removing internal representations, leaving underlying knowledge intact within the model.

\begin{figure}
    \centering
    \includegraphics[width=0.9\linewidth]{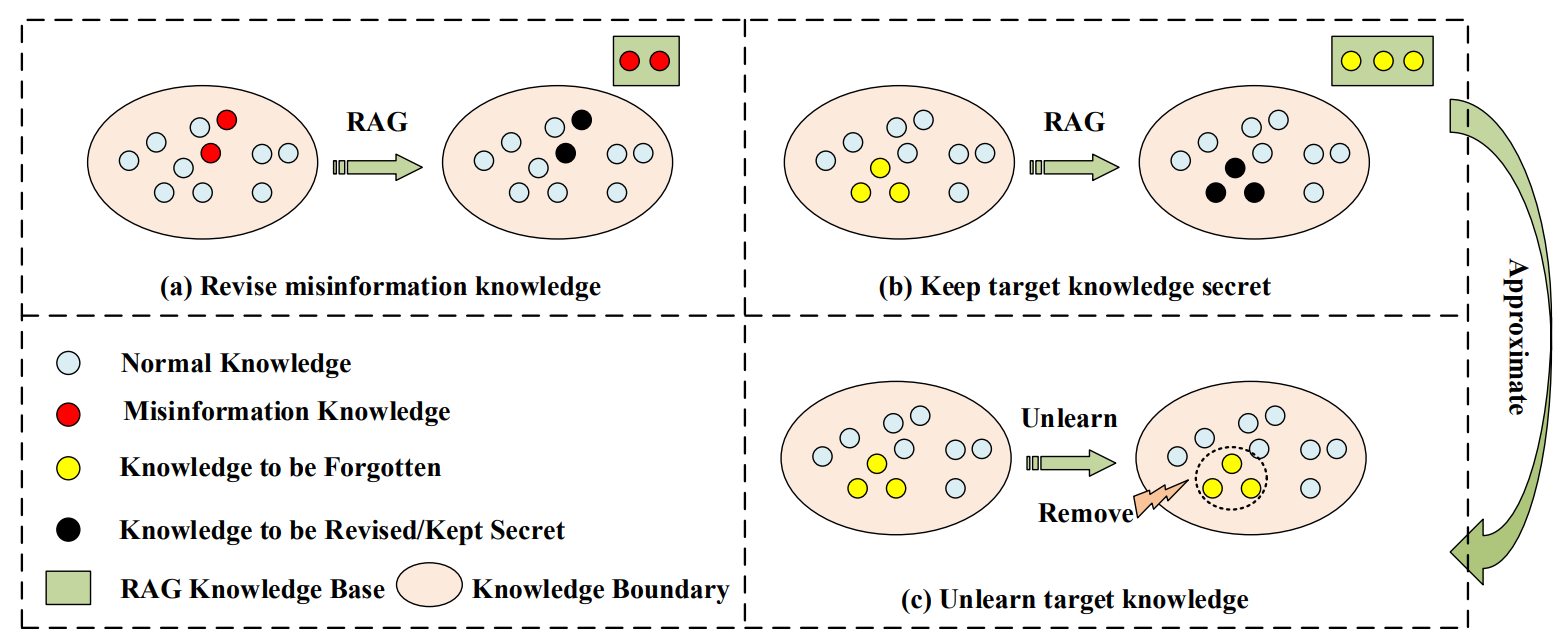}
    \caption{The intuition behind RAG-based unlearning \cite{wang2024machineunlearningmeetsretrievalaugmented}, showing three ways RAG-base systems can implement unlearning.}
    \label{fig:placeholder}
\end{figure}

Overall, other approaches highlight the breadth of perspectives on unlearning. By operating at the prompt or system level, they offer lightweight and often practical solutions, but they trade off strong guarantees of forgetting in favor of efficiency and reversibility.

\section{Unlearning Evaluation}
Assessing the effectiveness of unlearning is as critical as developing new methods. Without reliable evaluation, it is difficult to determine whether a model has truly forgotten targeted knowledge, whether its general utility has been preserved, or whether it remains vulnerable to adversarial extraction. Unlike standard model training, where accuracy or perplexity provide straightforward signals, unlearning requires specialized benchmarks and metrics that capture both \emph{forgetting} and \emph{retention}. At the same time, appropriate datasets are needed to probe model behavior across a wide range of tasks, domains, and languages.

In this section, we review existing evaluation practices along two complementary axes. First, we examine the \emph{benchmarks and metrics} proposed to quantify forgetting, utility preservation, efficiency, and robustness. Second, we survey the \emph{datasets} that serve as testbeds for unlearning, ranging from synthetic corpora to large-scale, real-world collections. Together, these resources form the foundation of the unlearning evaluation ecosystem and highlight the current gaps that must be addressed for the field to mature.

\subsection{Benchmarks and Metrics}
\label{benchmarks}
Evaluating unlearning requires specialized metrics that go beyond standard accuracy or perplexity. The key challenge is to measure not only whether the model has forgotten targeted knowledge, but also whether it continues to perform well on similar and unrelated knowledge.

\paragraph{Forget/Retain Evaluation.}  
Most unlearning methods for LLMs are evaluated at the fact or QA level, where the goal is to measure forgetting on a designated subset and retention on unrelated tasks. Forgetting effectiveness is typically assessed using multiple-choice questions, QA pairs, verbatim text reproduction, or fill-in-the-blank prompts associated with the forget and retain sets. Common benchmarks include TOFU \cite{maini2024tofutaskfictitiousunlearning} (memorized facts), WHP \cite{eldan2023whosharrypotterapproximate} (fictional QA), WMDP \cite{li2024wmdpbenchmarkmeasuringreducing} (biosecurity, cybersecurity, and chemical knowledge), and RWKU \cite{jin2024rwkubenchmarkingrealworldknowledge} (knowledge about well-known individuals).

\paragraph{Overlapping Domains}  
Recent work highlights that forget and retain sets are often constructed to be highly disjoint, which creates overly optimistic impressions of unlearning performance. In reality, there tends to be significant overlap between information that should be forgotten and information that should be preserved \citep{hu2025blurbenchmarkllmunlearning}. For example, a question about Botox as a migraine treatment involves benign medical knowledge (retain) and bioweapon misuse (forget) \citep{li2024wmdpbenchmarkmeasuringreducing}. To capture this ambiguity, a benchmark for LLM unlearning (BLUR; \citealp{hu2025blurbenchmarkllmunlearning}) constructs combined queries of forget questions and unrelated retain questions based on \citet{eldan2023whosharrypotterapproximate} and inserts forget-related terms into retain questions from \citet{li2024wmdpbenchmarkmeasuringreducing}. BLUR expands this evaluation to include combined queries, overlapping tasks, and relearning datasets of varying difficulty. Results show that existing methods often degrade sharply under these conditions, with simple baselines sometimes outperforming more sophisticated approaches. BLUR thus underscores the need for benchmarks that reflect real-world ambiguity and robustness to benign probing.

\begin{figure}[!b]
    \centering
    \includegraphics[width=0.9\linewidth]{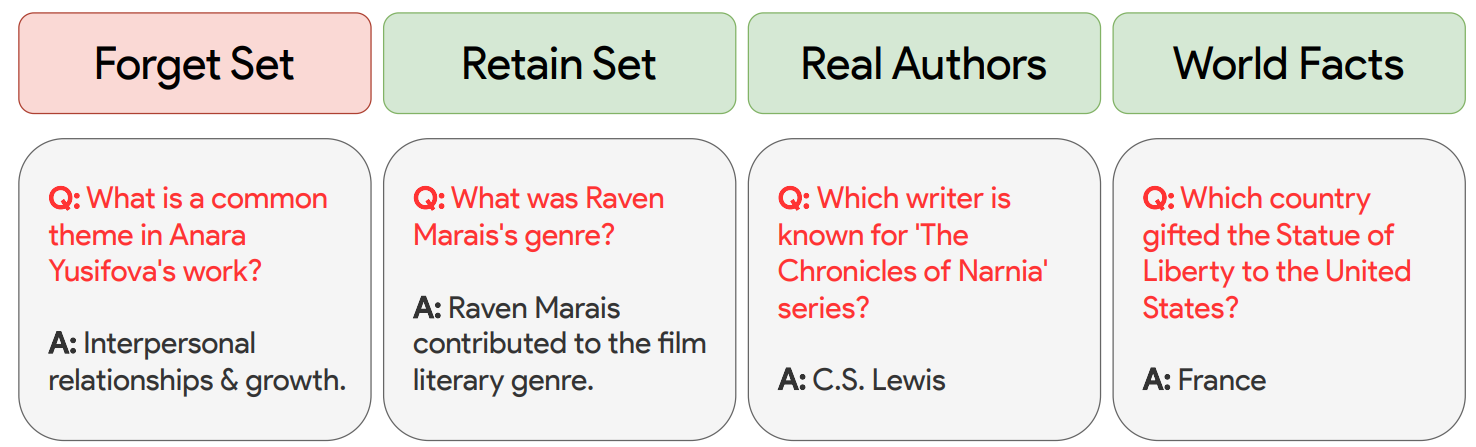}
    \caption{Example of forget and retain sets for unlearning evaluation in the TOFU dataset \cite{maini2024tofutaskfictitiousunlearning}}
    \label{fig:placeholder}
\end{figure}

\paragraph{Broad Domain Evaluation.}  
Beyond narrow forget/retain benchmarks, some approaches combine targeted and broad evaluations to better capture the trade-off between selective forgetting and overall utility. For instance, \textsc{UNLEARN} \citep{lizzo2025unlearn} evaluates forgetting on task-specific datasets such as NarrativeQA, Natural Questions, and GSM8K, while measuring retention across broader frameworks like HELM and Big-Bench. This dual evaluation highlights the necessity of testing both fine-grained forgetting and general performance preservation, especially as LLMs are expected to operate reliably across a wide range of tasks and domains.

\subsection{Datasets}
\label{datasets}

\paragraph{Synthetic Forgetting Sets.}  
Synthetic datasets provide an important avenue for evaluating LLM unlearning, offering controlled settings where forget and retain knowledge can be precisely constructed. Examples include TOFU’s synthetic fact memorization tasks \cite{maini2024tofutaskfictitiousunlearning}, synthetic PII or keyword-based corpora introduced in unlearning evaluations, and recent efforts that automatically generate domain-specific forget sets using LLMs themselves \cite{zhu2025llmunlearningexpertcurated}. Synthetic datasets are particularly valuable because they enable reproducibility and fine-grained control over what should be forgotten versus retained, though they may not always capture the complexity or ambiguity of real-world data distributions.

\paragraph{General LLM Datasets.}  
To assess knowledge retention and overall utility after unlearning, researchers rely heavily on broad evaluation suites designed to probe diverse model capabilities. The MMLU benchmark \citep{hendrycks2021measuringmassivemultitasklanguage} is among the most widely adopted, spanning 57 academic subjects and serving as a standard test of general knowledge retention. More recently, the HELM framework \citep{liang2023holisticevaluationlanguagemodels} has emphasized multidimensional evaluation—including accuracy, robustness, calibration, and fairness—making it especially valuable for studying trade-offs introduced by unlearning. Similarly, BIG-Bench \citep{srivastava2023imitationgamequantifyingextrapolating} provides an expansive collection of tasks created by over 400 researchers, covering reasoning, commonsense, and specialized domains. These broad datasets allow researchers to quantify the extent to which targeted forgetting affects unrelated capabilities, even though they were not originally designed with unlearning in mind.

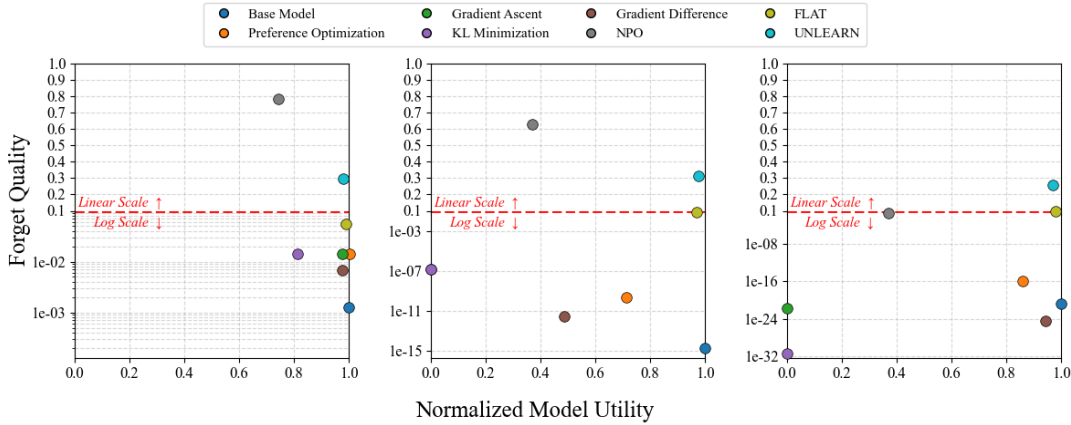
\begin{figure*}[!t]
    \centering
  \begin{overpic}[width=0.9\linewidth]{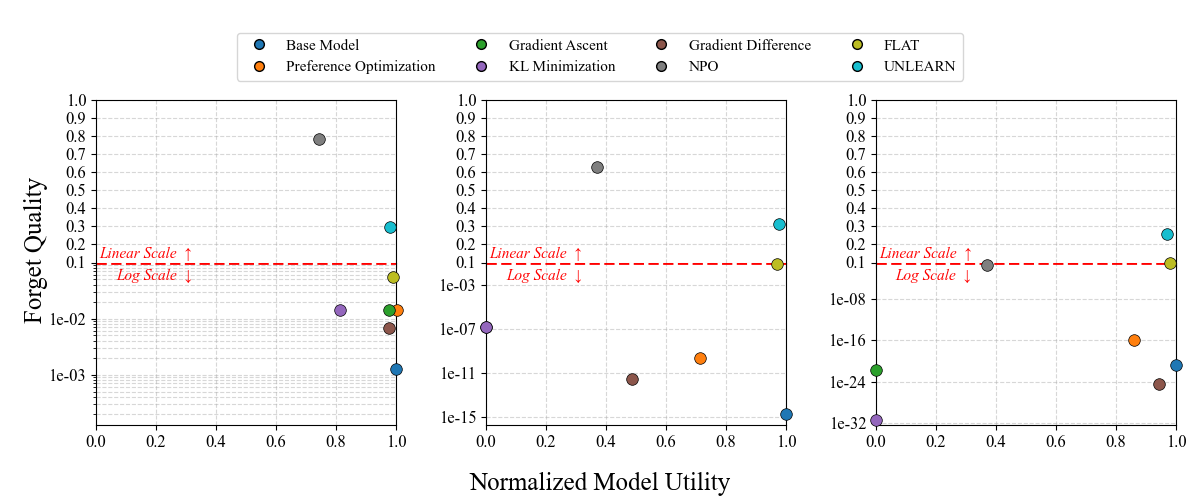}
    \put(23.5,35.7){\hyperlink{cite.rafailov2024directpreferenceoptimizationlanguage}{\tiny \phantom{XXXXXXXXXXXX}}}
    
    \put(42,37.4){ \hyperlink{cite.jang2023knowledge}{\tiny \phantom{XXXXXXXI}}}
    \put(42,35.7){ \hyperlink{cite.wang2025balancing}{\tiny \phantom{XXXXXXXX}}}

    \put(57,37.4){ \hyperlink{cite.liu2022continual}{\tiny \phantom{XXXXXXXXXI}}}
    \put(57,35.7){ \hyperlink{cite.wang2024flat}{\tiny \phantom{XX}}}

    \put(73,37.4){\hyperlink{cite.zhang2024npo}{\tiny \phantom{XXX}}}
    \put(73,35.7){ \hyperlink{cite.lizzo2025unlearn}{\tiny \phantom{XXXXXI}}}
  \end{overpic}
  \caption{Comparison of unlearning performance on the TOFU benchmark \cite{maini2024tofutaskfictitiousunlearning} across forget-set sizes (1\%, 5\%, 10\%). Each point represents a model's trade-off between normalized model utility (higher is better) and forget quality (p-value; higher is better).  The dashed red line marks the $\alpha = 0.1$ threshold distinguishing statistically significant forgetting (above the line) from unsuccessful forgetting (below the line), as well as a transition between linear and log scales for forget quality.}
    \label{fig:tofu_results}
\end{figure*}

\paragraph{Cross-Language and Multimodal Resources.}
Datasets for unlearning are beginning to extend beyond English text. In the multilingual setting, \citet{lu2025learnunlearnaddressingmisinformation} introduces multilingual QA and translation resources to test whether forgetting in one language transfers across languages, and \citet{choi2024crosslingualunlearningselectiveknowledge} expands cross-lingual unlearning by directly adding language-dependent weights for multilingual models. For multimodal unlearning, \citet{patil2025unlearningsensitiveinformationmultimodal} provides benchmarks for evaluating unlearning ability on the OK-VQA dataset. Similarly, SAFEERASER \citep{chen2025safeeraserenhancingsafetymultimodal} explores the multimodal space with a dataset of 3000 images and 29.8 VQA pairs. Together, these multilingual and multimodal resources remain less mature than text-only datasets, but they highlight the need to evaluate unlearning in more diverse settings.

Overall, datasets form the foundation of unlearning evaluation. Synthetic sets provide control, general benchmarks test retention, and cross-language or multimodal resources ensure broader applicability.

\subsection{Evaluation}
The TOFU benchmark \cite{maini2024tofutaskfictitiousunlearning} provides a controlled setting for evaluating unlearning through the construction of synthetic knowledge about imaginary authors and measuring a model's ability to selectively forget them. Unlike free-form probing or QA-style tests, TOFU offers more standardized forget and retain sets, enabling consistent comparison across methods. TOFU evaluates unlearning on two metrics:
\begin{itemize}
    \item Model Utility captures the model's retained capabilities after unlearning. The original TOFU paper used a harmonic mean of nine computed metrics on both the retain set, real authors, and world fact knowledge. This survey sticks with this convention; however, results are normalized against the base model. As such, a score of 1.0 indicates successful preservation of the retained knowledge after unlearning.
    \item Forget quality is expressed as a p-value comparing the distribution of the unlearned model's predictions on forget-set items with the distribution of the original model. When there are low p-values (below 0.1 in this study), there is a statistically significant deviation, indicating incomplete or unsuccessful forgetting. When there are p-values above 0.1, we cannot reject the null hypothesis that the two distributions are the same, indicating strong forgetting for any model above our $\alpha$ level.
\end{itemize}
Together, these two metrics capture the central trade-off in unlearning: effective removal of targeted knowledge versus preservation of a model’s broader capabilities. Figure~\ref{fig:tofu_results} visualizes this trade-off across a range of methods and forget-set sizes. Forget-set size refers to the proportion of synthetic author facts designated for removal (1\%, 5\%, or 10\% of the TOFU corpus), with larger proportions generally making the unlearning task more challenging.

At a glance, traditional unlearning approaches—such as preference optimization \cite{rafailov2024directpreferenceoptimizationlanguage}, gradient ascent \cite{jang2023knowledge}, KL minimization \cite{wang2025balancing}, and gradient difference methods \cite{liu2022continual}—consistently fall below the $\alpha = 0.1$ forget-quality threshold across all forget-set sizes; these approaches fail to produce strong forgetting.

More recent methods demonstrate improved performance but reveal different trade-offs. FLAT \cite{wang2024flat} achieves near-significant forget quality—just below the statistical threshold—while maintaining strong model utility. This suggests that the method strikes a partial balance between forgetting and capability preservation, though its ability to fully eliminate targeted knowledge remains limited. In contrast, NPO \cite{zhang2024npo} attains statistically significant forgetting across all forget-set sizes, indicating effective removal of targeted author knowledge. However, its utility sharply deteriorates as the forget-set size increases; there are significant, unintended effects on adjacent knowledge. 

Finally, \textsc{UNLEARN} \cite{lizzo2025unlearn} achieves statistically significant forgetting across all evaluated forget-set sizes while preserving normalized model utility near 1.0. This indicates that \textsc{UNLEARN} not only satisfies the forget-quality criterion but does so with minimal collateral degradation to retained capabilities. The separation between \textsc{UNLEARN} and other methods becomes more pronounced as the forget-set size increases, highlighting the importance of methods that explicitly isolate task-specific subspaces rather than relying on scalar penalty terms or direct gradient manipulation.

\section{Challenges and Open Problems}
Despite rapid progress, machine unlearning for large language models remains in its early stages. Current methods often trade off scalability, guarantees, or robustness, leaving significant gaps for future work. Below, we highlight several recurring challenges and open directions.

\paragraph{Guarantees of Forgetting.}  
A central question is how to provide verifiable guarantees that targeted information has been removed. Current evaluations rely on empirical probes: measuring performance drops or behavioral changes; however, these tests cannot ensure that forgotten data is unrecoverable \citep{jagielski2023measuringforgettingmemorizedtraining}. Formal guarantees remain elusive in large networks: differential privacy (DP) provides strong theoretical protection \citep{dwork2014} but these bounds degrade for over-parameterized models the size of LLMs. Recent work has begun to explore certifiable forgetting frameworks using statistical testing \citep{ginart2019makingaiforgetyou}, but none provide practical guarantees at LLM scale. Bridging the gap between the scale of LLMs with the mathematical rigor of DP remains a central challenge.

\begin{figure}[!b]
    \centering
    \includegraphics[width=0.9\linewidth]{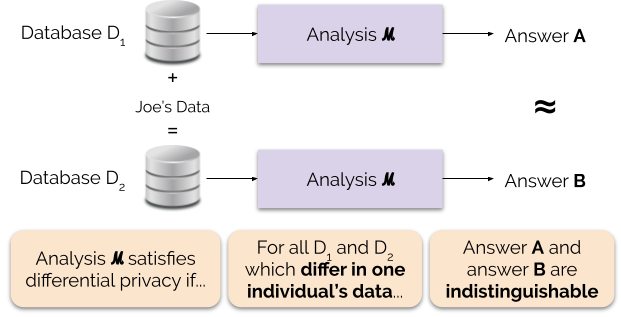}
    \caption{Differential privacy provides formal guarantees that model outputs remain indistinguishable when data is removed.}
    \label{fig:placeholder}
\end{figure}

\paragraph{Boundaries of Forgetting.}  
Even if unlearning reduces performance on a forget set, two key questions remain: did we erase all of the intended knowledge and did we retain everything else? These questions are compounded by the dichotomy between test sets and real-world data, where boundaries between data are as discrete as forget and retain sets. At the same time, excessive forgetting risks collateral degradation of unrelated tasks or domains \citep{lizzo2025unlearn}. Together these issues underscore the difficulty of defining precise boundaries between what must be erased and preserved. Formal mathematical tools that work at smaller scales have not been scaled and validated for reliable evaluation on LLMs.

\paragraph{Governance and Policy.}  
Unlearning is not only a technical challenge but also a matter of governance. Regulations such as GDPR and CCPA grant individuals the “right to be forgotten,” yet there is little consensus on how to operationalize this right in LLMs. Should forgetting be defined as empirical inability to reproduce training data, or as a formal guarantee \citep{ginart2019makingaiforgetyou}? This issue is compounded by the lack of accessibility to relevant data to create benchmarks to evaluate personalized unlearning. This highlights the need for standardized benchmarks, reporting practices, and compliance protocols to bridge technical research in machine unlearning with governance and audting frameworks.

\paragraph{Efficiency.}  
Retraining from scratch is infeasible for today's tens-of-billions-parameter models, and even partial retraining remains prohibilitively expensive at scale. Early works such as SISA training \citep{bourtoule2021machineunlearning} demonstrated the computational burden of data deletion by requiring repeated training on disjoint shards. More recent research has turned to adapter-based methods \citep{hu2021loralowrankadaptationlarge,gao2025largelanguagemodelcontinual} as a way to reduce cost. However, a persistent tension remains: lightweight approaches oten leave residual traces of the targeted knowledge, while methods with stronger guarantees typically demand signficantly more computation. Designing unlearning techniques that successfuly combine scalability, efficiency, and robust guarantees remains a central challenge for LLM unlearning.

\paragraph{Cross-Language and Cross-Modal Unlearning.}  
Multilingual unlearning introduces unique challenges the remain underexplored. Most existing studies focus on English, leaving open questions about how forgetting generalizes across languages with varying typologies, scripts, and resource availability. Initial efforts \citep{lu2025learnunlearnaddressingmisinformation} reveal that unlearning on information one language generally \emph{does not} lead to unlearning on other languages. Further, current multilingual benchmarks remain limited in scale and diversity \citep{choi2024crosslingualunlearningselectiveknowledge}, and systematic evaluation of over-forgetting, transfer, and intereference across langauges is lacking. Similar gaps existing in multimodal settings. While datasets like UnLOK-VQA \citep{patil2025unlearningsensitiveinformationmultimodal} and SAFEERASER \citep{chen2025safeeraserenhancingsafetymultimodal} provide early resources, multimodal unlearning remains far less studied than its text-only counterpart. Open problems include how to measure unleraning consistently across modalities, how to avoid unintended degradation across modalities, and how to construct sufficiently broad benchmarks beyond VQA-style tasks. These gaps underscore the need for broader resources for LLM unlearning to ensure methods are robust in diverse settings.

\paragraph{Adversarial Relearning.}  
Even if forgetting succeeds, forgotten information may re-emerge through fine-tuning, prompt engineering, or adversarial extraction. Recent work demonstrates that multimodal attacks can exploit residual representations in VLMs, enabling the recovery of ``forgotten" knowledge \citep{patil2025unlearningsensitiveinformationmultimodal}. Further works show the efficacy of such attacks: Stealthy Unlearning Attack \citep{zhang2025suastealthymultimodallarge} adds perturbation noise to ``sidestep" the forgetting process. These results highlight the shortcomings of existing unlearning methods against an adversary attempting to gain access to forgotten information.

Overall, these challenges highlight the gap between current empirical methods and the stronger guarantees, efficiency, and robustness needed for trustworthy unlearning at scale.

\section{Conclusion}

Machine unlearning has become an essential capability for large language models, driven by concerns of privacy, copyright, efficiency, and ethics. In this survey, we synthesized the expanding body of work on unlearning methods and organized them into data-centric, parameter-centric, architecture-centric, hybrid, and other approaches. We also examined the evaluation ecosystem, which includes benchmarks, metrics, and datasets designed to measure forgetting effectiveness, knowledge retention, and robustness. In addition, we identified persistent challenges such as establishing guarantees of forgetting, defining what knowledge should be erased or preserved, ensuring efficiency at scale, and extending unlearning across languages and modalities.

Together, these threads highlight both the promise of unlearning and the immaturity of current solutions. While initial results show that selective forgetting is possible, most existing methods involve trade-offs between practicality, completeness, and robustness. Progress will depend not only on advances in model architectures and optimization techniques, but also on the development of clear standards for evaluation, governance, and deployment. Our goal in this survey is to provide a roadmap that supports researchers and practitioners in building language technologies that are more trustworthy, efficient, and socially responsible.

\section*{Limitations}

\section*{Ethics Statement}

\section*{Acknowledgements}
This work was supported by CoCoSys, one of seven centers in JUMP 2.0, a Semiconductor Research Corporation (SRC) program sponsored by DARPA.

\bibliography{emnlp2023}
\bibliographystyle{acl_natbib}

\appendix

\end{document}